# MASNET: IMPROVE PERFORMANCE OF SIAMESE NETWORKS WITH MUTUAL-ATTENTION FOR REMOTE SENSING CHANGE DETECTION TASKS

Zhou Hongbin[1,2], Ren Yupeng[1,2,*], Li Qiankun[1], Yin Jun[1], Lin Yonggang[2]

[1] Zhejiang Dahua technology Co., Ltd, Hangzhou 310051, China
[2] Zhejiang University, Hangzhou 310058, China

**ABSTRACT:**

Siamese networks are widely used for remote sensing change detection tasks. A vanilla siamese network has two identical feature extraction branches which share weights, these two branches work independently and the feature maps are not fused until about to be sent to a decoder head. However we find that it is critical to exchange information between two feature extraction branches at early stage for change detection task. In this work we present **M**utual-**A**ttention **S**iamese **Net**work (MASNet), a general siamese network with mutual-attention plug-in, so to exchange information between the two feature extraction branches. We show that our modification improve the performance of siamese networks on multi change detection datasets, and it works for both convolutional neural network and visual transformer.

## 1. INTRODUCTION

Change detection is a challenging task in computer vision, whose goal is to assign pixel level binary label, change or non-change, to a pair of co-registered images. In the community of remote sensing, co-registered means two images taken at different time cover the exactly same geographical area. Along with abundant remote sensing images, change detection technology can be extremely useful in city planning, forest monitoring, disaster assessment, map update, etc.

Traditional change detection method can be divided into two categories, namely object-level change detection and pixel-level change detection, based on whether context information is used or not. Pixel-level methods are prone to noise as a single pixel does not carry much information. While object-level methods do not generalize form scene to scene as the hand-crafted feature extraction algorithm is not robust enough.

In recent years, deep learning gets more and more popular and brings performance boost in various fields of computer vision, and change detection is no exception. An intuitive way to handle change detection task with deep learning approach is to do it in a semantic segmentation manner. In (Peng, 2019) and (Jaturapitpornchai, 2019), the two co-registered images are stack together in channel-wise to be fused as a multi-band image, the fused image is processed by an encoder-decoder network subsequently, and a two-class segmentation mask is output as the change map. In such an approach, the fusion process is done before feature extraction, so this approach is also known as Early-Fusion. Early-Fusion forces the encoder to do feature extraction and change detection simultaneously, however, for most convolutional neural network based encoders, the receptive filed of convolutions in shallow layers is limited, so they have to detect change without enough context information. Even though theoretically feasible, context information lack in early stage may harm the performance of Early-Fusion network in practice.

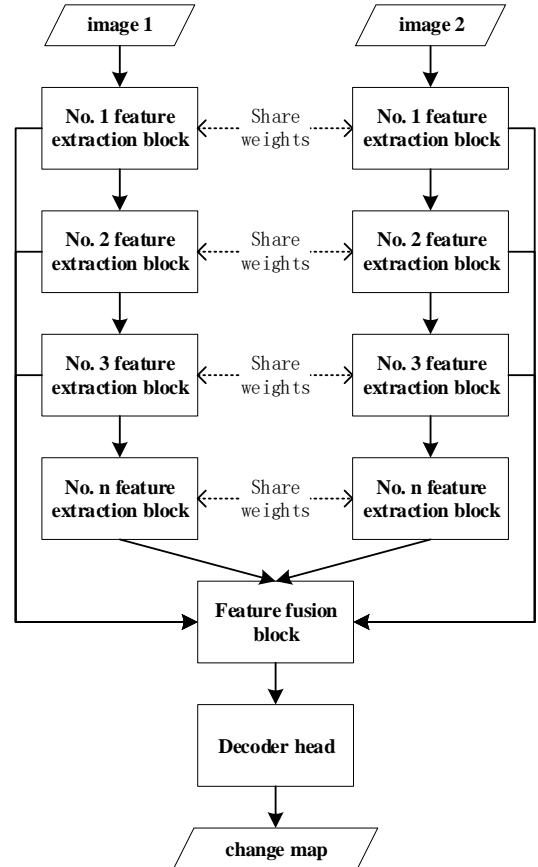

**Figure 1.** Structure of siamese networks.

Siamese network is first introduced in (Bromley, 1993), and in (Daudt, 2018) it is adopted as a kind of model specialized for change detection task, which adopts an encode-decoder architecture. A typical siamese network is shown in Figure 1, the type and number of feature extraction blocks may vary from design to design. The encoder contains two identical feature extraction branches which share weights. The co-registered images are processed by these two branches respectively, so that two feature maps are obtained independently. Then the two feature maps are fused to be one in the manner of stack channel-wise (Chen, 2020a), positional addition (Chen, 2021c) or positional difference (Chen, 2021a). The fused feature map is processed by a decoder head to produce the change map. Siamese network splits the feature extraction stage and change detection stage, so that the encoder can learn expressive feature maps and the decoder has global receptive field, which be of benefit to model performance.

* Correspondence: renyupeng@zju.edu.cn

Considering how we human beings do change detection ourselves, the main difference is at the feature extraction stage. There are abundant features on each image of co-registered pair, some features are relevant to change detection task, and some features are not, it's just depend on what features on another image. Whenever we human beings find a feature on one image, say a building, we will look it up on the other image: if there is a building in the same place on the other image, then we have to extract more features of the building, like building style or roof colour, to decide whether these two buildings are the same one; if there is no building in the same place on the other image, we know change happens, and we do not have to dig deeper for more features of the building. We human beings extract features on one image efficiently by peeking on the other image. Such information exchange between two images is critical for us to effectively filter out the most important features. However vanilla siamese networks can't exchange information between two images, as we have mentioned, that two feature extraction branches do their job independently. Indiscriminate feature extraction leads to less informative feature maps, which harms model performance of vanilla siamese networks.

In this work, we present MASNet, a siamese network with mutual-attention plug-in. The mutual-attention architecture sets up a bridge for the two feature extraction branches and helps them to tell task-relevant features from irrelevant ones. Experiment results show that mutual-attention makes a constant boost on various siamese networks and change detection datasets.

Our main contributions of this work are as follows: (1) we point out that information exchange in the feature extraction stage is critical for change detection task; (2) we present mutual-attention plug-in for siamese networks to make MASNet which improves model performance, our MASNet achieves SOTA results on LEVIR-CD dataset.

## 2. RELATED WORK

### 2.1 Attention Mechanism

Cross attention is a mechanism that reweight a feature vector V with the information from another feature vector Q. The attention mechanism is first introduced to neural machine translation (Bahdanau, 2014). When the vectors V and Q are from the same source, which means a vector reweights itself without extra information, it becomes a special form of attention called non-local (Wang, 2018) or self-attention (Vaswani, 2017), which is most used in deep learning community. Figure 2 exhibits a common implementation of self-attention, three vectors are obtained from the feature map by linear projection, then vector Q and K are fused by matrix multiplication followed with a softmax operation to get a weight vector, the weight vector is used to reweight vector V by matrix multiplication, and the output is added to the original feature map to form a residual learning mechanism (He, 2016).

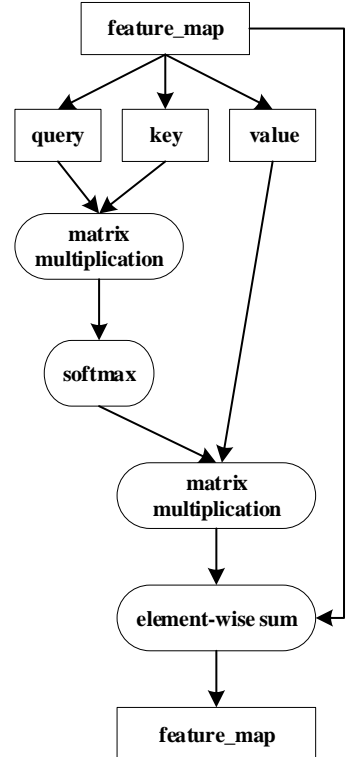

**Figure 2.** Structure of self-attention block.

Self-attention gains popularity in the field of natural language processing first, then this trend has spread to the field of computer vision. In (Fu, 2019) self-attention is introduced into fully convolutional neural network for semantic segmentation task, the self-attention is applied in channel-wise and spatial-wise to learn better feature maps. In (Dosovitskiy, 2020) an image classification model named Visual Transformer is built with self-attention based transformer block completely, rather than convolutional layer.

### 2.2 Siamese Network

To further improve the performance of siamese network, a bunch of modification ideas have been proposed. In (Chen, 2020b) spatial and channel attention is added to the encoder to capture long-range dependencies and to suppress pseudo-changes. In (Yang, 2020) siamese network with two different feature extraction branches is presented to deal with the problem that change area may have totally different land-cover distributions. In (Chen, 2020a) a spatial-temporal attention is presented to calculate attention weights between different times and different positions, which are used to get more discriminative feature maps. In (Chen, 2021c), (Daudt, 2019) and (Liu, 2020) an additional decoder is added to the model to constrain the feature extraction process, this additional decoder works as a semantic segmentation head and is trained with segmentation loss and/or self-supervised loss. In (Chen, 2021a) transformer is used in feature extraction branches for better long-rang dependencies in space-time. In (Chen, 2021b) a group point-wise convolution with channel shuffle is presented to make the siamese network very deep while efficient. In (Heidary, 2021) Gaussian attention is used in U-Net based siamese network to improve accuracy. These modification solutions may learn better feature extraction branches, however there is still no information exchange between two branches.

## 3. METHOD

As mentioned above, to find the most useful features for change detection on one image, we need some clue from another image. Inspired by attention and self-attention, we present mutual-attention block for siamese network, in which "mutual" means two feature maps reweight themselves with the information from each other.

As shown in Figure 3, *feature_map_1* and *feature_map_2* are the under-processing feature maps from *image_1* and *image_2* respectively, each feature map is linearly mapped to three vectors, namely value, key and query. To calculate the weight map of *feature_map_1*, we multiply *query_2* from *feature_map_2* with *key_1* from *feature_map_1*, so that *feature_map_1* can get a glimpse of what features are extracted in *feature_map_2*, with that information it can be learned which features should be strengthen and which features should be suppressed on *feature_map_1*. After a softmax operation, the weight map is used to reweight vector *value_1* by matrix multiplication, and the weighted result is added to the original *feature_map_1*. Above is how mutual-attention reweight *feature_map_1* with information from *feature_map_2*, and vice versa.

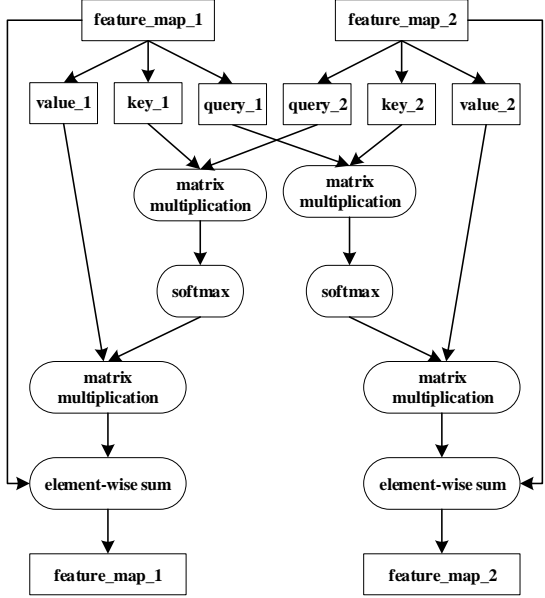

**Figure 3.** Structure of mutual-attention block.

The process in Figure 3 can also be expressed as the following equations, in which *x1* and *x2* are feature maps, *d* is the dimension of matrix *q1* and *q2*, *Wq*, *Wk* and *Wv* are learnable parameters that project feature map to *q*, *k* and *v*.

$$\begin{bmatrix} q1 & q2 \\ k1 & k2 \\ v1 & v2 \end{bmatrix} = \begin{bmatrix} Wq \\ Wk \\ Wv \end{bmatrix} \cdot [x1 \quad x2] \quad (1)$$

$$x1 = x1 + softmax\left(\frac{q2 \cdot k1^T}{\sqrt{d}}\right) \cdot v1 \quad (2)$$

$$x2 = x2 + softmax\left(\frac{q1 \cdot k2^T}{\sqrt{d}}\right) \cdot v2 \quad (3)$$

Figure 4 shows how we add mutual-attention to siamese network as a plug-in block to make **M**utual-**A**ttention **S**iamese **Net**work (MASNet). Most neural network backbones are designed as a stack of several feature extraction blocks ever since VGG (Simonyan, 2014), our solution is to add mutual-attention block follow every feature extraction block. We find such a way works well and will not increase too much computational cost.

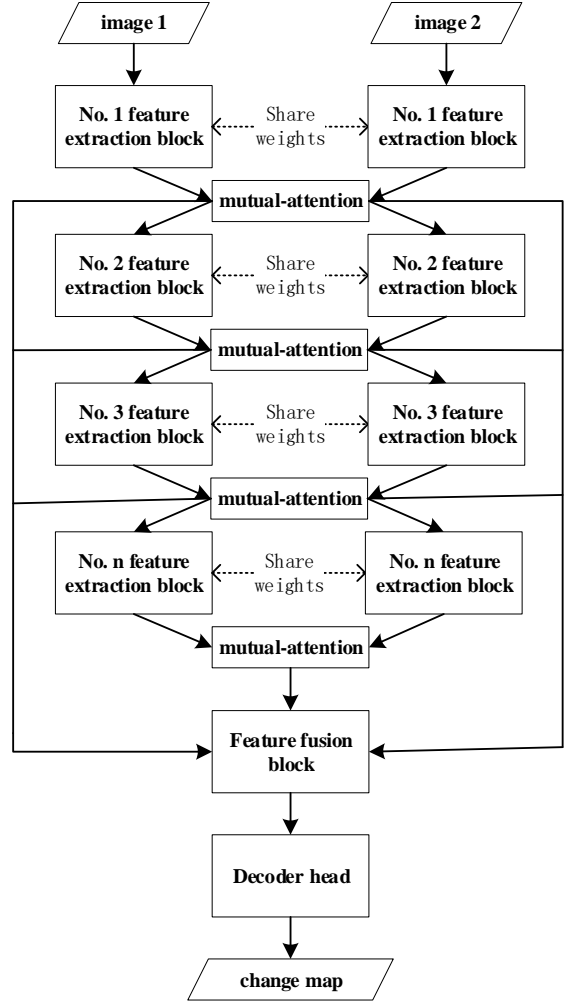

**Figure 4.** Structure of MASNet.

The feature map from a feature extraction block has the dimension of C*H*W, in which C for channel, H for height, and W for width. Generally speaking, C goes larger while H and W goes smaller as the feature extraction block goes deeper, which means more information and larger receptive field for each spatial position on the feature map. For two C*H*W feature maps, mutual-attention can be applied on three different level, namely global, local and individual. Global mutual-attention exchange information of the whole feature maps, which is intuitive. Local mutual-attention exchange information between two co-registered small windows, as shown in Figure 5, which has a dimension of C*h*w, h for window height and w for window width. Individual mutual-attention exchange information between two co-registered pixels, as shown in Figure 6, which has a dimension of C*1*1.

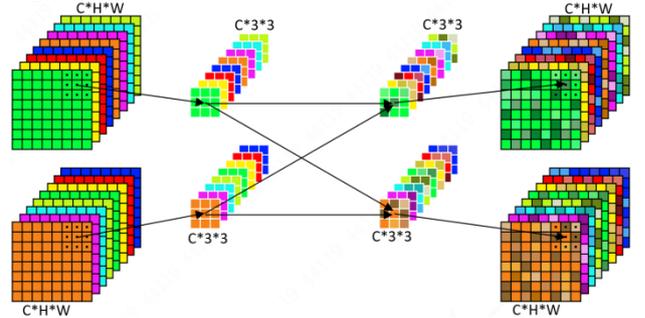

**Figure 5.** Local level Mutual-Attention with window of 3*3.

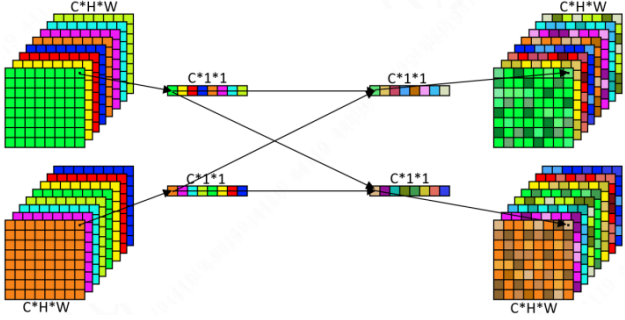

**Figure 6.** Individual level Mutual-Attention.

To limit the parameters and FLOPs increment caused by mutual-attention block, we apply individual level mutual-attention in our MASNet.

## 4. EXPERIMENT

### 4.1 Datasets

**LEVIR-CD dataset** is presented by (Chen, 2020a), it is specialized for building change detection. The co-registered remote sensing image pairs are collected from Google Earth, with resolution of 0.5 m/pixel. A total number of 637 pairs of images with a size of 1024*1024 pixels are collected in the area of Texas of the United States, seasonal change and illumination change are considered in data collection. The authors split the dataset into three part, namely training-set, validation-set and testing-set, in the ratio of 7:1:2. The dataset is open-sourced on https://justchenhao.github.io/LEVIR/. In Figure 7 are some data samples of the dataset.

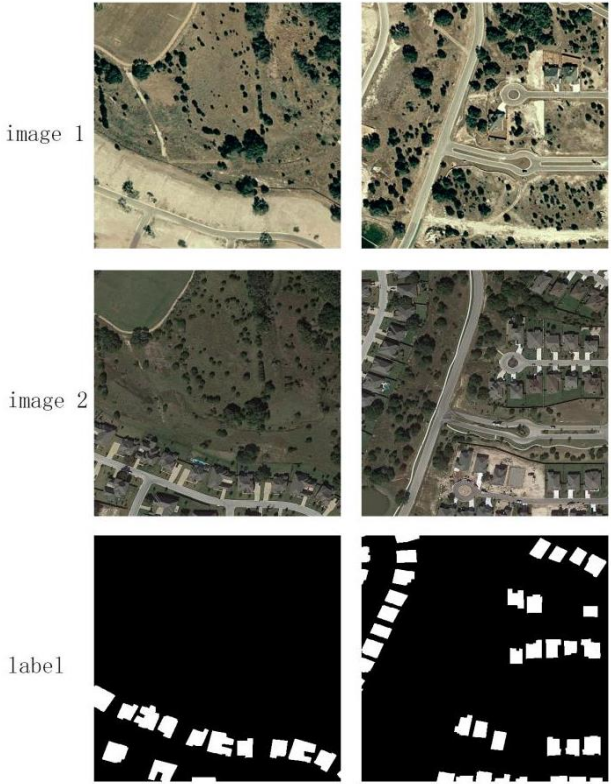

**Figure 7.** Data sample of LEVIR-CD dataset.

**SECOND dataset** is presented by (Yang, 2020), it is specialized for semantic change detection, which assigns an additional semantic label to each changed pixel. The dataset focuses on 6 semantic classes, namely non-vegetated ground surface, tree, low vegetation, water, buildings and playgrounds. The data is collected from cities of Hangzhou, Chengdu and Shanghai, a total number of 4662 pairs are collected with a size of 512*512 pixels, in which 2968 pairs belongs to training-set and 1694 pairs to testing-set. The dataset is open-sourced on http://www.captain-whu.com/PROJECT/SCD/. By ignoring the semantic label, we convert the SECOND dataset to a standard binary change detection dataset, as shown in Figure 8.

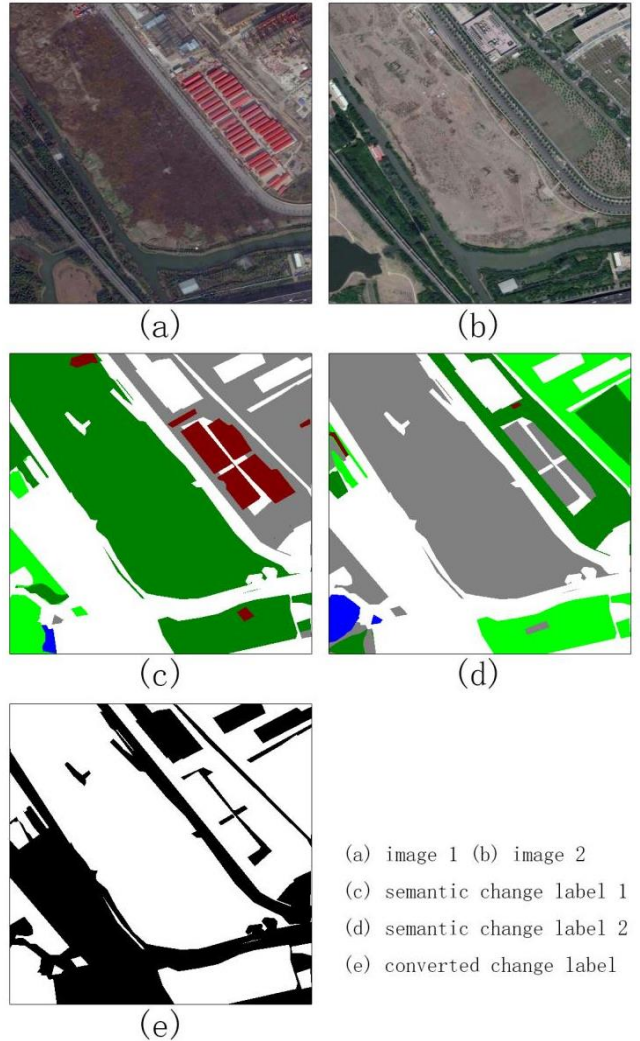

**Figure 8.** Data sample of SECOND dataset.

### 4.2 Evaluation Metrics

We use IoU of change class as the metric to compare MASNet and vanilla siamese network models. When evaluate our MASNet on LEVIR-CD, the F1 score is added as it is used by many other researchers. These two metrics are calculated as follows:

$$IoU = \frac{TP}{TP + FN + FP} \quad (4)$$

$$F1 = \frac{TP}{TP * 2 + FN + FP} \quad (5)$$

In which, *TP* refers to true positive, which is the number of pixels that belongs to change area in ground truth and predicted to be change; *FN* refers to false negative, which is the number of pixels that belongs to change area in ground truth but predicted to be non-change; *FP* refers to false positive, which is the number of pixels that belongs to non-change area in ground truth but predicted to be change.

**4.3 Experiment Settings**

We test our mutual-attention plug-in on two models, namely HRNet-OCR (Sun, 2019) and Segformer (Xie, 2021), one for convolutional neural network and one for visual transformer, two most popular kinds of models in computer vision.

The backbone of HRNet has 4 stages. For vanilla HRNet-OCR based Siamese network, we stack the output of stage 4 of two feature extraction branches in channel-wise, then a point-wise convolution layer (Lin, 2013) is applied to recover the channel dimension. For MASNet model, we apply mutual-attention plug-in right after each stage, and the feature fusion method is the same with vanilla approach. The 10 mutual-attention blocks, one for each resolution path in each stage, added to MASNet model of HRNet-OCR brings 1.82% more parameters compared to the vanilla one.

The backbone of Segformer also has 4 stages, and all of their outputs are used by the decoder head, so that the channel-wise stack feature fusion is applied on all 4 stages. And mutual-attention plug-in is applied right after each stage. The 4 mutual-attention blocks added to MASNet model of Segformer brings 2.38% more parameters compared to the vanilla one.

For SECOND dataset, because the labels of testing-set is not available to us, so we take 4-folds cross validation on the training-set. The images in training-set are named by numbers, so we sort the images by name and split them evenly into 4 subsets. A model is trained on 3 subsets and test on the rest one, the mean score of the 4 subsets is adopted to evaluate this model. As 4-folds cross validation is robust to random inference, so we don't train a model several times.

For LEVIR-CD dataset, we follow the original dataset split, the model is trained on training-set and we save checkpoints regularly, and the checkpoint which performs best on validation-set is chosen and tested on testing-set. To reduce random inference, we train each model 3 times with random seed and report the mean score.

The segformer is initialized with an official b4 checkpoint, while HRNet with an official w48 checkpoint, both pre-trained on ImageNet. The training super-parameters are shared by vanilla models and MASNet models, the images are resize with a random scale between 0.5 and 2.0, the resized images are randomly cropped to 512*512, flip and rotate augmentation are applied, before sent to network, these two augmented images are randomly switched. AdamW (Loshchilov, 2019) is applied as optimizer with initial learning rate of 6e-5 and weight decay of 0.01, and the learning rate is scheduled with poly policy with power of 1.0, we warm up the learning rate in the first 1500 iterations from 0 to initial learning rate. RMI loss (Zhao, 2019) is used to supervise the training process. The models are trained on 8 NVIDIA 1080ti GPUS, the batch size is set to 16, and the max iteration is set to 80000, for each 8000 iterations a checkpoint is saved.

**4.4 Experiment Results**

Overall, the presented MASNet models outperform vanilla ones, for two different kinds of models on two different datasets.

As shown in Table 1, mutual-attention brings constant performance boost on SECOND dataset. For HRNet-OCR, the mean score of 4 folds is increased from 53.70% to 55.59%, a gain of 1.89 is achieved. While for Segformer, the mean score of 4 folds is increased from 51.73% to 55.27%, a considerable gain of 3.54 is achieved. On each and all folds, MASNet models out-perform vanilla siamese network models, which shows the effectiveness and robustness of MASNet.

| models | fold1 | fold2 | fold3 | fold4 | mIoU |
|---|---|---|---|---|---|
| HRNet-OCR vanilla | 49.57 | 57.32 | 54.91 | 53.00 | 53.70 |
| HRNet-OCR MASNet | 51.91 | 58.71 | 56.55 | 55.18 | **55.59** |
| Segformer vanilla | 47.51 | 54.81 | 52.14 | 52.45 | 51.73 |
| Segformer MASNet | 51.52 | 58.29 | 56.18 | 55.34 | **55.27** |

**Table 1**. Models performance (IoU, %) on SECOND dataset.

From statistic in Table 2 we can see that mutual-attention brings noticeable performance boost compared to vanilla siamese network, on LEVIR-CD dataset. For HRNet-OCR with mutual-attention, the IoU score on testing-set is improved from 85.45% to 85.79%, an average gain of 0.34 is achieved. And for Segformer with mutual-attention, the IoU score is improved from 84.32% to 85.00%, an average gain of 0.68 is achieved. The improvement is small compare to SECOND dataset, we think it is because LEVIR-CD only concerns building appearance and disappearance, which is less challenging for models and so has less space for improvement, while the SECOND dataset has change type between 6 classes. It is worth mention that the worst model in 3 repeated tests of MASNet still outperforms the best model in 3 repeated tests of vanilla siamese network, which shows the robustness of MASNet.

| models | validation | test |
|---|---|---|
| HRNet-OCR vanilla | 86.04±0.13 | 85.45±0.07 |
| HRNet-OCR MASNet | 86.41±0.05 | **85.79±0.07** |
| Segformer vanilla | 84.62±0.03 | 84.32±0.11 |
| Segformer MASNet | 85.68±0.05 | **85.00±0.10** |

**Table 2**. Models performance (IoU, %) on LEVIR-CD dataset.

As an open dataset, many researchers use LEVIR-CD to test their models, the F1 score and IoU score of these models are reported in Table 3. Some works that did not report IoU score are marked with "--". It shows that our HRNet-OCR model with mutual-attention has a F1 score of 92.62% and an IoU score of 86.26% on LEVIR-CD testing-set, both achieving SOTA performance. Our best model is selected from the 3 repeated tests as mentioned above, and inference with extra TTA, in which we set the inference multi-scale to be [0.5, 0.75, 1.0, 1.25, 1.5, 1.75, 2.0].

| methods | F1 score | IoU score |
|---|---|---|
| FCN+PAM (Chen, 2020a) | 87.3 | -- |
| CANet (Lu, 2021) | 87.4 | -- |
| Multibranch (Zhao, 2021) | 88.04 | -- |
| FODA (Zhang, 2021) | 88.73 | -- |
| MCCRNet (Ke, 2021) | 90.71 | -- |
| ChangeStar (Zhen, 2021) | 91.25 | 83.92 |
| CEECNet V2 (Diakogiannis, 2021) | 91.83 | 84.89 |
| FCCDN (Chen, 2021c) | 92.29 | 85.69 |
| MASNet (ours) | **92.62** | **86.26** |

**Table 3**. Models comparison on LEVIR-CD dataset.

In Figure 9 we try to check if MASNet works as we design by visualizing some mutual-attention maps, the first row contains original image 1 and mutual-attention maps which are applied to the feature maps of image 1 on each stage of HRNet, and the second row contains original image 2 and corresponding mutual-attention maps. Information exchange can be observed in the mutual-attention maps. From the column of stage 1, we can see that the mutual-attention map contains the low level features from another image. As the network goes deeper, the mutual-attention maps contain higher level features, like buildings in the column of stage 3, and it is worth note that the attention map highlight the pixels of feature map of image 1 where there is building in image 2, and vice versa. In the column of stage 4, the mutual-attention map suppress hardly the pixels of new building on image1, while highlight the pixels of new building on image 2, and both attention maps put a mid-tone to the non-change buildings.

Above phenomena hints that the structure of MASNet has the ability to exchange information at feature extraction stage, which improves the model performance on change detection tasks.

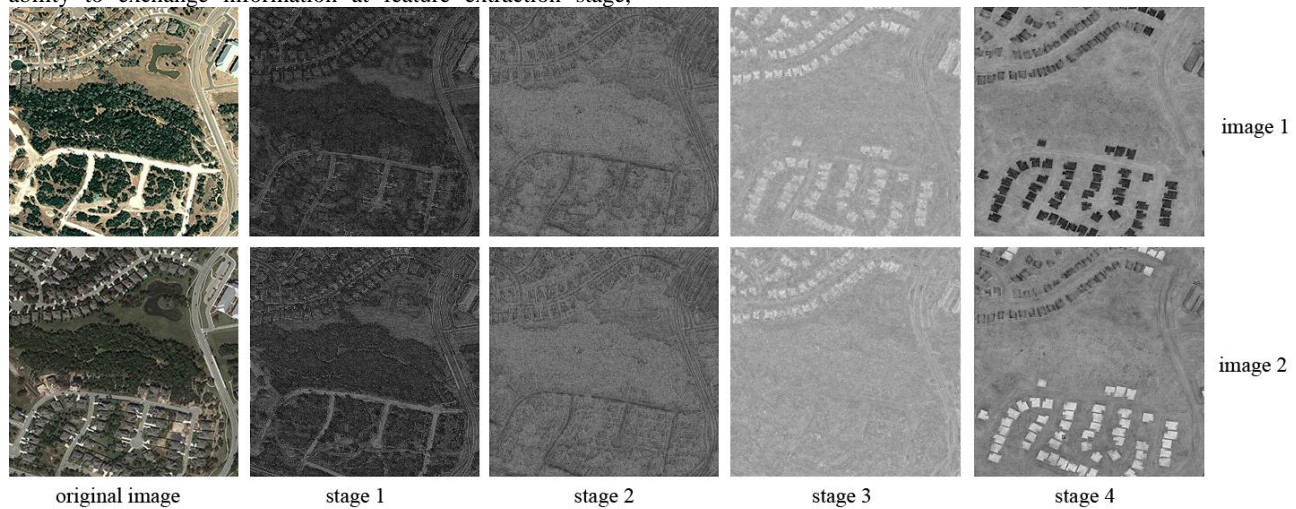

**Figure 9**. Visualization of mutual-attention maps

## 5. CONCLUSIONS

In this work we propose the idea that information exchange in feature extraction stage is critical for siamese network to improve their performance on change detection tasks. Without such information exchange, vanilla siamese networks have to extract all features, relevant or irrelevant, to make the change map, which is less optimal. We present a plug-in named mutual-attention, which can be added to any siamese network to make MASNet, to achieve the purpose of information exchange. Experiment results show that mutual-attention works for different kind of models on various datasets. Our best MASNet model achieves SOTA performance on LEVIR-CD dataset. We believe that mutual-attention is not the best way to exchange information, more work should be done to design better structure in the future.